\documentclass[sigconf]{acmart}

\usepackage{booktabs} % For formal tables
\usepackage{listings}
\usepackage{balance}
\usepackage{fancyhdr}
\usepackage{multirow}
\usepackage{enumitem}
\usepackage{ wasysym }

\begin{document}

\fancyhead{}
\balance
\sloppy

\copyrightyear{2019} 
\acmYear{2019} 
\setcopyright{acmcopyright}
\acmConference[WSDM '19]{The Twelfth ACM International Conference on Web Search and Data Mining}{February 11--15, 2019}{Melbourne, VIC, Australia}
\acmBooktitle{The Twelfth ACM International Conference on Web Search and Data Mining (WSDM '19), February 11--15, 2019, Melbourne, VIC, Australia}
\acmPrice{15.00}
\acmDOI{10.1145/3289600.3291018}
\acmISBN{978-1-4503-5940-5/19/02}

\lstset{basicstyle=\footnotesize\ttfamily}

\author{Christoph Hube and Besnik Fetahu}
 \affiliation{
   \begin{tabular}{c}
	 L3S Research Center, Leibniz University of Hannover\\
     Hannover, Germany \\
    \{hube, fetahu\}@L3S.de
   \end{tabular}
}

\title{Neural Based Statement Classification for Biased Language}
	
\begin{abstract}

Biased language commonly occurs around topics which are of controversial nature, thus, stirring disagreement between the different involved parties of a discussion. This is due to the fact that for language and its use, specifically, the understanding and use of phrases, the stances are cohesive within the particular groups. However, such cohesiveness does not hold across groups.

In collaborative environments or environments where impartial language is desired (e.g. Wikipedia, news media), statements and the language therein should represent equally the involved parties and be neutrally phrased. Biased language is introduced through the presence of inflammatory words or phrases, or statements that may be incorrect or one-sided, thus violating such consensus.

In this work, we focus on the specific case of phrasing bias, which may be introduced through specific inflammatory words or phrases in a statement. For this purpose, we propose an approach that relies on a recurrent neural networks in order to capture the inter-dependencies between words in a phrase that introduced bias. 

We perform a thorough experimental evaluation, where we show the advantages of a neural based approach over competitors that rely on word lexicons and other hand-crafted features in detecting biased language. We are able to distinguish biased statements with a precision of $P=0.917$, thus significantly outperforming baseline models with an improvement of over 30\%.  Finally, we release the largest corpus of statements annotated for biased language.
\end{abstract}

%\keywords{Biased Language; Wikipedia Quality; NPOV}

\maketitle

\section{Introduction}\label{sec:introduction}

In sociolinguistic theory, language and its linguistic structures are seen as a medium that is in the function of specific \emph{social groups}~\cite{halliday1970language}. That is, language and its use reflects the demands and other characteristics of the group (i.e., ideology, economical, cultural). This usually results in a consensus amongst a group in the vocabulary use and the meaning of specific phrases and words on specific topics. Due to the diversity in stances and points of view for different topics, such a consensus cannot always be achieved. This is often the case when written discourse (or any language utterance) is considered to be biased. Bias in language is manifested in different forms, from discrimination in terms of \emph{gender} through over-lexicalization (e.g. \emph{female doctor} vs. \emph{doctor} for male)~\cite{romaine2000language}, or in terms of \emph{authority} on how one addresses a person (e.g. \emph{nominal} reference through \emph{title + name} vs. name)~\cite{brown1960pronouns,fowler2013language}. Other forms of bias tackle the believability of a statement or introduce terms that are considered to be one-sided in topics that do not have a consensus amongst the different societal groups~\cite{recasens2013linguistic}.

Wikipedia is a unique environment in manifesting such diversity in terms of points of view and stances for a large variety of topics. The current English version of Wikipedia consists of more than 5 million articles of highly diverse topics, which are constructed from a large editor base of more than 32 million editors\footnote{\url{https://en.wikipedia.org/wiki/Wikipedia:Statistics}}. Given its scale and diversity it is not surprising that many statements in Wikipedia reflect biases from its underlying editors, respectively their societal background. Statements on issues that are controversial cause disagreements between editors, specifically the different points of view in a discussion. Other factors are diffused from external sources like news, the second most cited external resource~\cite{DBLP:conf/websci/FetahuAA15,DBLP:conf/cikm/FetahuMNA16}. Fowler~\cite{fowler2013language} shows that news are prone to a range of issues such as language bias.

To avoid such cases of language bias and other biases that arise in controversial topics, Wikipedia has established a set of principles and guidelines. For instance, the \emph{neutral point of view} (NPOV) defines criteria that should be followed by its editors: (i) avoid stating opinions as facts, (ii) avoid stating seriously contested assertions as facts, (iii) avoid stating facts as opinions, (iv) prefer nonjudgemental language, and (v) indicate the relative prominence of opposing views. Recent work \cite{recasens2013linguistic,martin2017persistent} shows that in Wikipedia's case\footnote{The author of the work in \cite{martin2017persistent} carried out an experimental study on his personal Wikipedia page \url{https://en.wikipedia.org/wiki/Brian_Martin_(social_scientist)}}, most NPOV violations are w.r.t biased language (i) -- (iv), and often one-sided statements (v), specifically in the form of \emph{epistemological} and \emph{framing} bias. \emph{Epistemological} refers to linguistic cues that have impact in the \emph{believability} of a statement, while \emph{framing} refers to the terms and phrases that are one-sided in the case where a topic may have multiple viewpoints.

The statements below show the diverse forms of bias that are present in Wikipedia.

\begin{enumerate}[leftmargin=*,label=(\alph*)]
	\item Andrew James Breitbart was one of the \emph{\textbf{most outspoken}}, \emph{\textbf{fearless}} conservative journalists in America.	
	\item The Labour Party in the United Kingdom put together a \emph{\textbf{highly successful}} set of policies based on encouraging the market economy, while promoting the involvement of private industry in delivering public services.

	\item An abortion is the \emph{\textbf{murder}} of a human baby embryo or fetus from the uterus, resulting in or caused by its death.
	\item Sanders \emph{\textbf{shocked}} his fellow liberals by putting up a Soviet Union flag in his Senate office.

	\item This may be a result of the fact that the public had \emph{\textbf{unsurprisingly}} lost support for the President and his policies.
	\item The Blair government had \emph{\textbf{promised}} a referendum on whether Britain should sign the Constitution, but \emph{\textbf{refused}} popular demands that it carry out its promise.
\end{enumerate}

\noindent The examples above show different forms of biased language. The cases in (a) -- (b) represent framing bias and are manifested in the form of adjectives like \emph{highly successful} or \emph{fearless} as subject intensifiers. The remaining cases represent epistemological bias, e.g., \emph{shocked} states a very strong precondition of the truth of the proposition (i.e., \emph{``shocking his fellow liberals''}), similar is (f).

In this work, for the mentioned aspects, we focus on detecting biased language in Wikipedia statements that are introduced either due to inflammatory wording or phrases and whether a statement is written in a neutral tone, thus, following the principles of the NPOV policy. Contrary to previous work which has addressed partially the problem of detecting biased language \cite{recasens2013linguistic,hube2018detecting}, where feature-based models are used to capture the different forms of bias in Wikipedia statements based on specific lexicons and hand-crafted features. However, such approaches fail to capture the inter-dependency of words that may incur bias and their context, and furthermore, relying solely on hand-crafted lexicons has its disadvantages of not being able to capture all forms of bias.

We propose an approach that is based on recurrent neural networks (RNN) with two modes of attention, \emph{global} and \emph{hierarchical}~\cite{bahdanau2014neural,yang2016hierarchical}, which achieves significant improvements over feature-based approaches \cite{recasens2013linguistic,hube2018detecting}. 

To this end, we provide the following contributions:
\begin{itemize}
	\item a neural model for detecting biased language,
	\item largest corpus of biased statements
\end{itemize}

\section{Related Work} \label{sec:related_work}

In this section, we review related work, which covers several aspects of \emph{biased language} and other forms of linguistic manifestation of biases, such as \emph{framing analysis} or \emph{gender biases}. In terms of corpora, most of related work is focused on Wikipedia, news media, and other political corpora like political debates. In the following, we categorize the related work based on their objective.

\textbf{Article Bias.} The seminal work by Greenstein and Zhu~\cite{greenstein2012wikipedia} is the first to analyze bias in Wikipedia. They adapt an approach initially developed for determining newspaper slant \cite{gentzkow2010drives}. The approach relies on a list of 1000 terms and phrases typically used by either republican or democratic congress members. To measure political bias in Wikipedia, Greenstein and Zhu look for occurrences of these terms and phrases in Wikipedia articles about US politics. For example, if an article contains significantly more terms typically used by democratic congress members compared to terms typically used by republican congress members, then this is an indicator for a pro-democratic leaning of the article's content. According to their findings, Wikipedia articles are on average more left-leaning, especially in the early phase of Wikipedia. With more editors working on an article, the bias decreases on average. But since most articles do not receive much attention, there is still a significant number of articles containing bias.

In their work, \cite{greenstein2012wikipedia} focus on the topic of \emph{US politics}. Therefore the framing bias that they detect has a narrow scope, whereas our work is different in the sense that we aim at capturing a broader scope of biased language. We classify statements that contain biased language which is introduced through words or phrases that are partial or are not neutrally phrased.

\textbf{Biased Language.} Recasens et al. \cite{recasens2013linguistic} propose an approach for detecting a single bias-inducing word given a biased Wikipedia statement. The approach relies on linguistic features, divided into two bias classes: framing bias, including subjective language such as praising and perspective-specific words; and epistemological bias, dealing with believability of a proposition, i.e. phrasing choices that either cast doubt on a fact or try to sell an opinion as a fact. In their dataset collection, they crawl Wikipedia revisions that have a ``\emph{POV}'' flag in the revision comments. We use a similar dataset collection procedure, however, we additionally use crowdsourcing to filter statements that do not contain bias ($>$ 60\% for our data sample). Given that their approach is originally intended to identify words that introduce bias, we adopt their approach and consider the proposed features in ~\cite{recasens2013linguistic} to classify statements as either containing bias or not as one of our baselines.

An additional competitor to our work is the work by Hube and Fetahu~\cite{hube2018detecting}. Their task is similar to ours, where they gather biased statements from Wikipedia by extracting statements from the right-wing conservative wiki Conservapedia\footnote{\url{http://www.conservapedia.com}}. Their approach extends on the work of \cite{recasens2013linguistic} by introducing features that take into account the context of certain words from lexicons, and additionally including features based on the LIWC text analysis tool~\cite{pennebaker2001linguistic}. The approach is supervised and one of their best features is a biased word list they construct by identifying bias word clusters in the Wikipedia word2vec word embedding space. We compare against this work, and show that long-range dependencies between words and phrases in a statement are hard to capture through hand-crafted features. 

Our Neural Network based approach shows significant improvement over the competitors from \cite{hube2018detecting,recasens2013linguistic}.

\textbf{Ideological Bias and Framing Analysis.} Iyyer et al. \cite{iyyer2014political} introduce a RNN model for classifying statements as either liberal or conservative. Their datasets contain statements from politicians in US Congressional floor debates and statements from ideological books about US politics. For pre-selecting biased statements from the data they make use of the features used by Yano et al. \cite{yano2010shedding} and a simple classifier with manually selected partisan unigrams as features. For labeling the pre-selected statements they use crowdsourcing, where crowdworkers label not only the full statement but also each phrase part of the sentence separately in a hiearchical manner. These additional labels allow for a handling of semantic compositions and the correct classification of more complex sentence structures, when the sentence parts are incrementally combined. For example, the statement \emph{They dubbed it the ``death tax'' and created a big lie about its adverse effects on small businesses.} is classified as liberal bias, even though the term ``death tax'' suggests pro-conservative bias. 

Lahoti et al.\cite{lahoti2017Joint} propose an unsupervised approach for determining the ideology of both users and content in a combined liberal-conservative latent space using Twitter data. They include features such as the surrounding network structure of users and information about content shared by users.

Baumer et al.~\cite{baumer2015testing} propose a model to detect the linguistic cues that introduce framing in political events. The results suggest that readership is often unaware of the subtle framing cue words, and that depending on the framing of an event the perception and stances towards an event may vary. The classifier relies on a set of syntactic and lexical features for identifying framing cue words. Similar is the work by Tsur et al.~\cite{tsur2015frame}, where they propose a topic modeling approach to identify farming words in news articles.

Our work is not comparable to the above works. The works in \emph{ideological} bias can be seen as a case of framing bias, whereas in the case of \emph{framing analysis}, the problem is even more subtle than framing bias. Framing usually represents the interplay between the cognitive bias and the context in which a statement is positioned. As such, the scope of these works cannot capture all the possible cases that we tackle and that introduce biased language.

\textbf{Other Bias.} Some research also covers other types of bias, e.g. selection bias \cite{bourgeois2018selection}, \cite{martin2017persistent} and bias focusing on specific topics, such as gender bias \cite{wagner2015s} or cultural bias \cite{callahan2011cultural}. We do not consider open opinions to be bias. For example, the statement \emph{I think this movie is really bad} is not bias according to our definition because the writer makes clear that it is her own opinion.

\section{Data Collection}\label{sec:data_collection}

In this section we introduce our approach on collecting statements from Wikipedia articles that contain biased phrasing. The data collection consists of two main procedures: (i) a pre-selection of statements from Wikipedia revisions that contain a \emph{POV} tag in the comments, and (ii) a crowdsourcing step which we use to manually annotate statements containing phrasing bias. Below we describe in details the individual steps.

\subsection{Extracting POV-tagged Statements from Wikipedia} \label{sec:pov-tagged}

Wikipedia editors are encouraged to add comments when changing or adding content in a Wikipedia article. In some cases editors add comments to mark that their change aims at reducing bias and thus restoring the \emph{Neutral Point of View}. 

We extract all statements from the entire revision history of the English Wikipedia, for those revisions that contain the \emph{POV} tag in the comments. This leaves us with 1,226,959 revisions. We compare each revision with the previous revision of the same article and filter revisions where only a single statement has been modified\footnote{With modified we understand any statement that has been \emph{updated/deleted/moved.}} The reason for this is that if multiple statements have been modified, we are unable to say if the POV tag in the revision comment refers to all statements or only to a fraction. The final resulting dataset leaves us with 280,538 \emph{pov-tagged} statements.

Table \ref{table:edit_types} shows the number of different edit types. In 129,578 cases the statement has been deleted in the new revision. In 601 cases the statement has been moved to a different section.

\noindent In another 150,359 cases, the statement has been updated in the new revision\footnote{We use assume that a statement has been updated if there is another statement that is similar to the previous statement with a high jaccard similarity of 0.7.} The low number of moved statements is not surprising, since moving a statement to another section does usually not mitigate its bias.

\begin{table}[ht!]
\centering
\smallskip
\begin{tabular}{ c c c }
\toprule
 \emph{deleted} & \emph{moved} & \emph{updated}  \\
 \midrule
 129,578 & 601 & 150,359\\
 \bottomrule
\end{tabular}
\caption{\small{Statements from revisions with \emph{POV} comments across the different modification types they undergo.}}
\label{table:edit_types}
\end{table}
\vspace{-10mm}

\subsection{Crowdsourced Ground-Truth Construction} \label{sec:crowd}

Wikipedia is a highly dynamic platform. Its user base is very large, and with it there is a high diversity in the expertise, that is, understanding the NPOV principle of Wikipedia, or simply there may be different stances towards an added statement in a Wikipedia page representing some form of event. We notice several additional types of biases that cause disagreement between the different Wikipedia editors as indicated by their revision comments:

\begin{itemize}
\item \textbf{Selection Bias:} \emph{"NPOV; the CS Monitor accusations are not relevant here"}
\item \textbf{Focus Bias:} \emph{"Actually, this info is already in the criticisms section. While I agree it is needed in the article multiple mentions is POV pushing."}
\end{itemize}

In other cases editors use the POV-tag to discuss the article's assumed bias:
\begin{itemize}
\item \emph{"can someone explain to me what is POV about this article?"}
\end{itemize}

Even in cases where the editor explicitly tags the statement as containing (phrasing) bias, this still reflects the opinion of only one editor. Other editors might disagree. 

\begin{figure*}[ht!]
	\centering
	\frame{\includegraphics[width=0.8\textwidth]{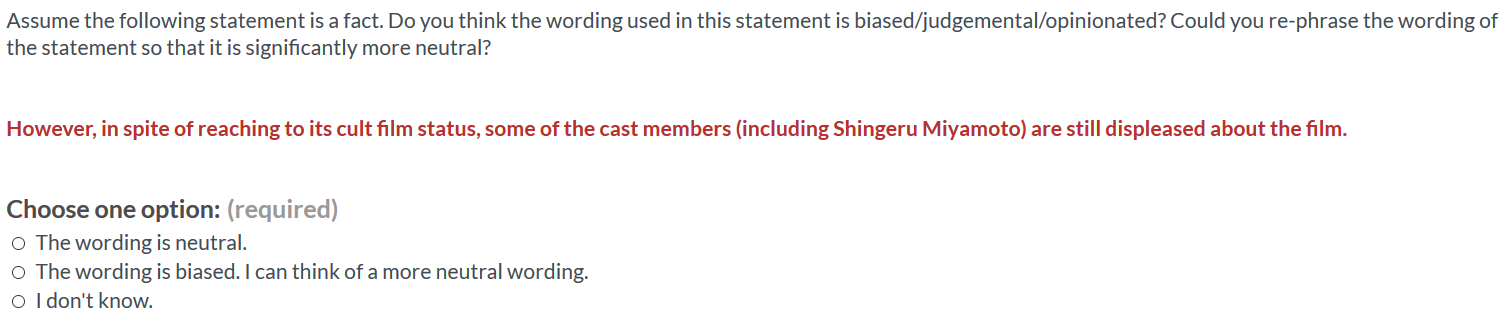}}
	\caption{Crowdsourcing job setup for annotating sentences as \emph{``biased''} or \emph{``neutral''}.}
	\label{fig:crowdsourcing_preview}
\end{figure*}

\paragraph{\textbf{Crowdsourced Ground-truth.}}
To tackle these issues, we ask workers to identify statements containing phrasing bias in the Figure Eight platform\footnote{\url{https://www.figure-eight.com/}}. Since labeling the full \emph{pov-tagged} dataset would be too expensive, we take a random sample of 5000 statement from the dataset. Figure \ref{fig:crowdsourcing_preview} shows a preview of the job, where we show a single statement to the workers and let them label each statement, providing three options: 

\begin{itemize}
\item \emph{``The wording is neutral.''}
\item \emph{``The wording is biased. I can think of a more neutral wording.''}
\item \emph{``I don't know.''}
\end{itemize}

Workers were allowed to choose only one option. Note that we are not just asking the workers to label statements according to whether they contain (phrasing) bias or not, since this would be a more ambiguous and subjective task. Instead we ask workers to consider the statement as a fact and to choose the \emph{``biased''} option only if they can think of a more neutral wording to present this fact. This way we make sure that the workers focus on the phrasing of the statement and not on it's content.

To improve quality of the judgments we provide a number of examples and restrict workers to level 2 or higher\footnote{Figure Eight divides workers into 3 levels with increasing competence.}. Additionally we set in place unambiguous test questions and filter out workers who do not pass at least 70\% of the questions.

For each judgment we pay \cent1.6 US cents. For each statement we collect 3 judgments leading to a total of 15,000 judgments. We measure worker agreement using Krippendorffs Alpha, a measure of rater agreement for sparse cases where not every rater rates every item. The agreement is low ($\alpha$ = 0.124) as expected given the subjectivity of the task.

We filter out all statements labeled as \emph{``I don't know''} and all statements with confidence $<0.6$. The final dataset contains 4952 labeled statements with 1843 ($\sim$ 37\%) of them labeled as biased and 3109 ($\sim$ 62\%) labeled as neutral. The large percentage of statements not labeled as biased confirms that the crowdsourcing step is necessary to identify the statements containing phrasing bias. Simply assuming that a POV-tagged statement contains phrasing bias would result in a larger but also low quality dataset.

\section{Biased Language Classification}\label{sec:approach}

In this section, we present our approach for classifying biased language in Wikipedia statements. We overcome some of the major drawbacks of feature based models in \cite{recasens2013linguistic,hube2018detecting}, which rely on hand-crafted features and specific lexicons, and thus, are limited in capturing the varying manifestations of bias in language.

As the following examples show, the mere presence of words cannot be considered to be a reliable indicator of bias. The first case shows a biased statements, whereas the second refers to an objective legal term. In addition, in other cases the bias can be introduced through phrases or multiple words appearing in different locations in a sentence (cf. third example below.).

\begin{itemize}
\item An abortion is the \emph{murder} of a human baby embryo or fetus from the uterus, resulting in or caused by its death.
\item In 2008 he was convicted of \emph{murder}.
\item \emph{The public agrees} that it is the \emph{number one} country in the world.
\end{itemize}

We remedy all of the above issues of existing work and propose two sequence based classifiers that rely on Recurrent Neural Networks (RNNs) with gated recurrent units (GRU)~\cite{cho2014learning} for computing the hidden representation of sequences in a sentence. Additionally, we will make heavy use of attention mechanisms~\cite{bahdanau2014neural,yang2016hierarchical} to determine words in a sentence that are indicators of biased language. We first describe the means with which we represent statements, then describe the necessary details of RNN, and finally explain in detail the two proposed models for biased language classification.   

\subsection{Statement Representation}

An important prerequisite in successfully applying RNN models in our task, is the representation of words in a sentence. We distinguish three main sentence representations.

\textbf{Word Representation.} We represent a sentence $s=(w_1,\ldots, w_n)$ consisting from a sequence of words through their corresponding word representations. We will use the GloVe embeddings~\cite{pennington2014glove} to represent the words in our corpus. Unknown words we will initialize randomly in our word embedding matrix $W_{glove}$.

Word embeddings have been successfully applied in downstream tasks in NLP, and are shown to be efficient in capturing context and synonymous words. 

\textbf{POS Tags.} POS tags are one of the most basic features used to represent text, and are able to capture stylistic linguistic features. POS tag are successfully employed in determining \emph{text genre}~\cite{biber1991variation}. Similarly, POS tags have shown to provide insights in determining biased statements in \cite{hube2018detecting}.

We additionally represent each token in $s$ through its POS tag. In our RNN models, we compute the POS tag embedding matrix $W_{POS}$, and use it in combination with $W_{glove}$.
 
\textbf{LIWC Word Functions.} LIWC text analysis~\cite{pennebaker2001linguistic} has been successfully employed in a number of tasks that capture subjectivity of text, such as analyzing language in \emph{fake news}~\cite{rashkin2017truth}, and additionally as shown in \cite{hube2018detecting}, LIWC features when used together with the context of the $n$--grams proves to provide a high improvement over existing approaches~\cite{recasens2013linguistic} in detecting biased statements.

Similarly as for POS tags, here too we train our embedding matrix $W_{LIWC}$ and use it in combination with other token representations. LIWC categorizes words into 75 different categories, each representing the function of a word, e.g. whether a word represents negative emotion. Since a word may be in function of different LIWC categories, we chose the most \emph{descriptive LIWC category} for a word\footnote{We compute an IDF measure on the word - LIWC function association, thus, we prefer LIWC function that are less likely to be assigned to other words.}. In general, LIWC categories express a range of psychological and sociological functions of words, and thus, are highly important for subjective tasks like detecting statements with biased language.

\subsection{RNN Statement Encoding}\label{subsec:representation}

For a given Wikipedia statement which we represent as a sequence of words $s = (w_1, \ldots, w_n)$, RNNs encode the individual words into a hidden state $h_t=f(w_t, h_{t-1})$. The function $f$ in our case can be represented either through an LSTM or GRU function\footnote{A detailed description of LSTMs and GRUs is beyond the scope of this work, we refer to the respective papers for more details~\cite{hochreiter1997long,cho2014learning}.}.

The encoding of an input sequence from $s$ is dependent on the previous hidden state. This dependency based on $f$ determines how much information from the previous hidden state is passed onto $h_t$. For instance, in case of GRUs, $h_t$ is encoded as following:
\begin{equation}
	h_t = (1 - z_t) \odot h_{t-1} + z_t \odot \tilde{h}_t
\end{equation}
where, the function $z_t$ and $\tilde{h}_t$ are computed as following:
\begin{align}
	z_t = \sigma( W_z w_t + U_z h_{t-1} + b_z) \\
	\tilde{h}_t = \tanh\left(W_h w_t + r_t \odot (U_h h_{t-1} + b_h)\right) \\
	r_t = \sigma( W_r w_t + U_r h_{t-1} + b_r)
\end{align}

The function $z_t$ decides the amount of information that is kept from $h_{t-1}$, which is extracted from step $t-1$ and thus impacts the computation of $h_t$, whereas $r_t$ is known as the \emph{reset gate} that can disallow the past information from $h_{t-1}$ to be included in $\tilde{h}_t$, which consequentially impacts the computation of the state $h_t$. 
This particular property of RNN encoders is highly important for our task, as the presence of words or phrases in statements with biased language can be easily encoded through the hidden states $h_t$. Furthermore, sequences which do not contribute in improving the classification accuracy are captured through the model parameters in function $r_t$, allowing for the model to ignore information coming from such sequences.

\subsection{RNN -- Global Attention }

One disadvantage of plain RNN models is that when used for classification tasks or language generation (standard encoder-decoder cases) tasks, the classification is done based on the last hidden state $h_N$. In the case of long sentences, this can be problematic as the hidden states, respectively the weights from the different input sequences have to be correctly represented in the last state.

Attention mechanisms~\cite{bahdanau2014neural} have proven to be successful in circumventing this problem. The main application of attention mechanism has been applied in machine translation tasks~\cite{bahdanau2014neural,luong2015effective}. The main difference between standard training of RNN models, is that all the hidden states are taken into account to derive a \emph{context vector}, where different states contribute with varying weights, or known with \emph{attention weights} in generating such a vector. The context vector, depending on the task, for instance in machine translation it is used to decode the input sequence into another sequence.

\begin{figure}[h!]
	\centering
	\includegraphics[width=1.0\columnwidth]{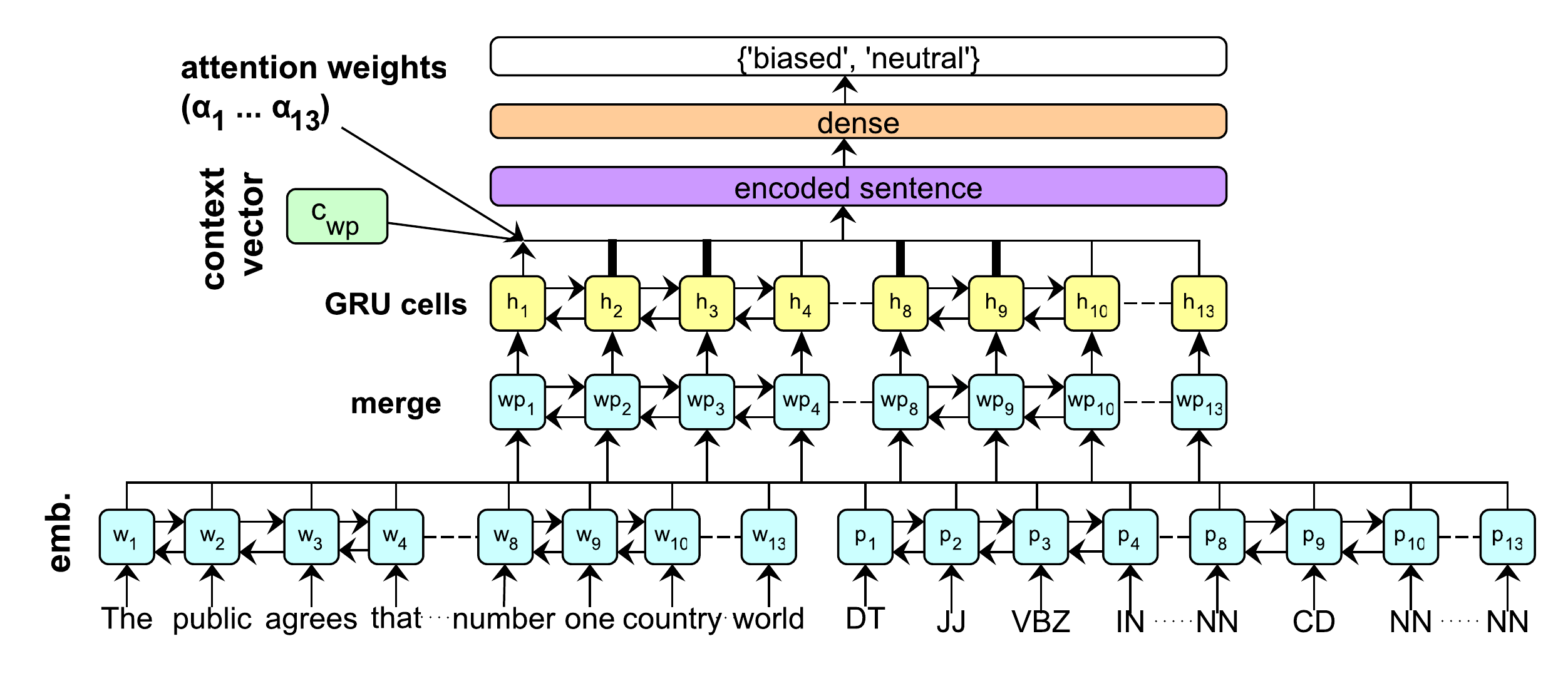}
	\caption{\small{We combine the different sentence representations by concatenating them. We compute a sentence representation based on a \emph{attention-mechanism}, which weighs the input sequences and thus generating the sentence representation based on their importance in the classification task.}}
	\label{fig:model_1}
\end{figure}

In our case, as shown in Figure~\ref{fig:model_1}, we employ the attention mechanism to compute a sentence representation $s_{rep}$ and use it to classify $s$. This has the advantage as our sentence representation consists only of the hidden states which are important in determining the class of $s$. More formally, we compute $s_{rep}$ as following:
\begin{align}
	u_t = \tanh\left(W_{emb}h_t + b_{emb}\right) \\
	\alpha_t = \frac{\exp(u_{t}^{T}c)}{\sum_{t'}{\exp(u_{t'}^{T}c})} \\ 
	s_{rep} = \sum_{t}\alpha_t h_t
\end{align}
We see from Eq (7) that $s_{rep}$ is the sum of the hidden states of $s$ weighted according to the importance of each sequence $\alpha_t$, where $\alpha_t$ simply represents a \emph{softmax} function over the hidden representation of words as computed in $u_t$ and the context vector $c$.

Finally, to account for different representations of $s$ such that we capture aspects such as the stylistic and LIWC features (see Section~\ref{subsec:representation}). We consider different combinations in our experimental setup, i.e., words + POS, words + LIWC, and words + POS + LIWC. We \emph{concatenate} the different sequence representations (see \emph{merge} layer in Figure~\ref{fig:model_1}), and pass them onto the GRU cells for learning the hidden representations $h_t$.

\subsection{RNN -- Hierarchical Attention}

Hierarchical attention, introduced in \cite{yang2016hierarchical}, is employed in the case of document classification. It first applies the attention mechanism on top of sentences, respectively at the word level. The computed word attention is used to represent a sentence, similar as in $s_{rep}$ for which they compute the hidden representations through GRU cells. Finally on top of the hidden representation of individual sentences is applied the attention mechanism, thus, resulting in a final document representation, which is used for classification.

\begin{figure}[ht!]
	\centering
	\includegraphics[width=1.0\columnwidth]{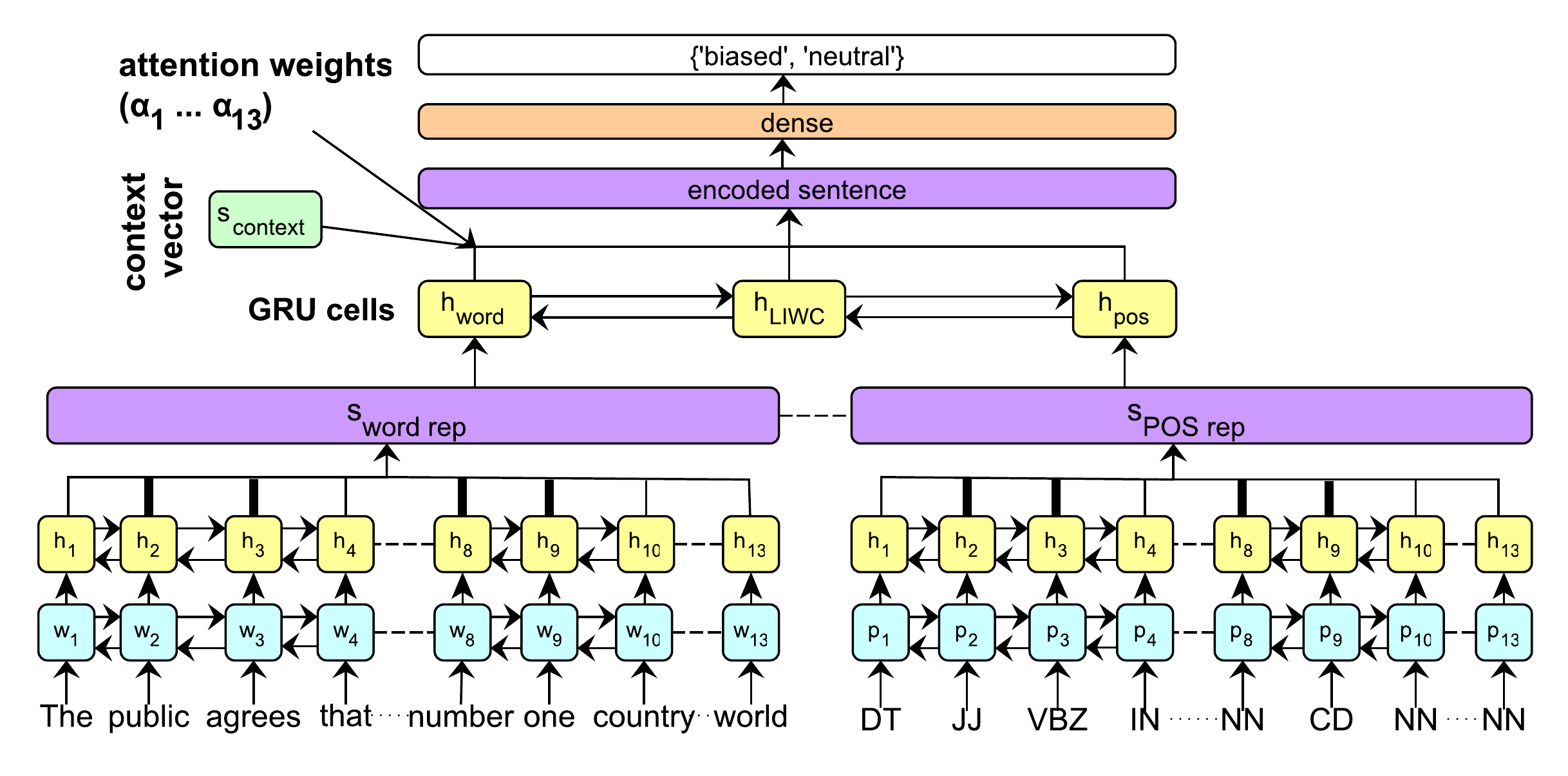}
	\caption{\small{We compute separately the attention weights and the corresponding sentence representations similar to Eq (7) for the different sentence representations. We pass the the computed sentence representations into GRU cells, thus, computing their hidden representations, from which we compute another joint representation based on the attention weights of the separate sentences, and finally classify using a \emph{sigmoid} function into \emph{``biased''} or \emph{``unbiased''}.}}
	\label{fig:model_2}
\vspace{-5mm}
\end{figure}

Here, we employ a similar strategy, in that we have a fixed set of sentence representations (see Section~\ref{subsec:representation}), which we feed as separate sentences into the hierarchical attention mechanism, and thus, are able to learn separately the importance of the different representations in determining if a sentence has biased language or not. Figure~\ref{fig:model_2} shows an overview of the proposed model. The computation of the overall sentence representation is similar to that in Eq (7). The only difference here lies in the fact that instead of merging the different sentence representations, we compute individually the importance of each representation. 

\section{Experimental Setup}\label{sec:setup}

In this section we describe the experimental setup for detecting statements that contain biased phrasing. We first describe the different strategies on generating datasets with \emph{unbiased} statements (apart from the ones gathered through crowdsourcing) and further describe the competitors and the learning setup of our approach.

\subsection{Datasets}\label{sec:datasets}

The statements that were marked as \emph{``neutral''} by crowdworkers in our data collection in Section~\ref{sec:data_collection} represent statements that contain other forms of biases or subjectivity as explained earlier (i.e., selection, focus biases etc.). As such these statements do not represent the ideal high quality content in Wikipedia. For this reason, we will denote the crowdsourced neutral statements as the \emph{hard} case of distinguishing between biased and neutral statements.

To obtain a cleaner labeled dataset containing both statements with and without biased phrasing, we additionally extract statements from featured Wikipedia articles which arguably contain mainly statements without biased phrasing due to their high quality. In the following, we describe all different datasets that we use for evaluating our approach.

\paragraph{\textbf{CW-Hard:}} This dataset consists of only the crowdsourced statements that we described in Section~\ref{sec:data_collection}. The dataset consists of 1843 statements marked as \emph{``biased''} and 3109 marked as \emph{``neutral''}. As we will see later in the evaluation section that this dataset proves to be the hardest as the \emph{``neutral''} statements contain quality issues that can be attributed to other forms of bias or subjectivity factors. 
	
\paragraph{\textbf{Featured:}} To extract \emph{``neutral''} statements of high quality, we turn back to statements extracted from featured articles in Wikipedia\footnote{Featured articles are considered to be articles of high quality conforming to the various editing policies in Wikipedia, such as: neutrality, statements that are verifiable through citations and additionally with highly reputable citations.}. Featured articles are more likely to be neutral when compared to statements from random articles of varying quality. The findings are consistent with~\cite{greenstein2012wikipedia}, where articles with a large number of revisions from a diverse pool of editors are less likely to contain bias. 

The English Wikipedia contains 5338 articles that are featured articles. We crawl the content of featured articles\footnote{Time of access: June 14th 2018} and extract more than 1 million statements, from which we sample the equivalent amount of statements (1.843 statements) as for the \emph{``biased''} class in our data collection step in Section~\ref{sec:data_collection}. Finally, the \emph{``biased''} statements in the \emph{\textbf{featured}} dataset are the same as in \emph{\textbf{cw-hard}}, with the only difference in the \emph{``neutral''} statements.

\paragraph{\textbf{Type--Balanced:}} Statements we extract in Section~\ref{sec:data_collection} are from a wide range of types of articles. Depending on their type (i.e. the Wikipedia categories an article belongs to or the type from a reference knowledge base), the statements therein will vary in their language genre and linguistic structure due to the difference in focus. For instance, articles about location vary substantially from articles about persons in their genre and topical focus in the respective articles.

Table \ref{tab:type_distributions} shows the top--$10$ types for the \emph{\textbf{cw-hard}} dataset and the \emph{\textbf{featured}} articles datasets. The type distributions in both datasets are different. While the \emph{\textbf{cw-hard}} dataset contains a larger number of articles belonging to the types \texttt{Place} and \texttt{Populated Place}, the \emph{\textbf{featured}} dataset contains mostly articles belonging to types like \texttt{Software}, \texttt{VideoGame}, and \texttt{MusicalWork}.

To account for such divergence in statement distribution in the \emph{\textbf{featured}} dataset, we enforce that statements should be from featured articles and additionally have a similar type distribution as the \emph{\textbf{cw-hard}} dataset. As we will show in the evaluation results later on, statements differ significantly across types in their thematic aspects and in some cases in language genre. Again, we take a random sample of 1.843 statements, similar to the amount of \emph{``biased''} statements as in the \emph{\textbf{cw-hard}} dataset. 

\begin{table}[ht!]
\centering
\smallskip
\begin{tabular}{ c c c }
\toprule
 \textbf{Type--Balanced} & \textbf{CW-Hard} & \textbf{Featured}  \\
 \midrule
Agent & Place & Work \\
Work & PopulatedPlace & Agent \\
Place & Agent & Software \\
Person & Settlement & VideoGame \\
PopulatedPlace & Organisation & Organisation \\
Organisation & Work & MusicalWork \\
Settlement & Country & Film \\
Species & Person & Album \\
Eukaryote & Company & Place \\
WrittenWork & City & Person\\
\bottomrule
\end{tabular}
\caption{\small{Top 10 Wikipedia article types from DBpedia for \emph{type--balanced} featured articles, \emph{cw-hard}, and \emph{featured} articles.}}
\label{tab:type_distributions}
\end{table}\vspace{-25pt}

\subsection{Baselines}\label{sec:baselines}

We compare our approach against two existing baselines, which focus on the same task as ours. The approaches rely on hand-crafted features to detect biased language in Wikipedia statements. Additionally, we consider as baselines vanilla RNNs without attention for varying sentence representations.

\begin{itemize}[leftmargin=*]
	\item \textbf{B1:} The first baseline is an adoption of the approach in ~\cite{recasens2013linguistic}. Originally the approach detects words that introduce biased statements. We adopt it such that instead of classifying individual words, we classify statements as either biased or not. The feature space is the same as in the original paper in \cite{recasens2013linguistic}.
	\item \textbf{B2:} The second baseline is a recent work which shows improvement over \textbf{B1}. The approach in \cite{hube2018detecting} extends over \textbf{B1} by further introducing contextual features by means of \emph{n-grams} and other features that analyze statements for psychological and sociological insights through the LIWC~\cite{pennebaker2001linguistic} text analysis tool.
	\item \textbf{RNN:} We consider as a baseline vanilla RNNs, where we compute the hidden representation of sequences with GRUs~\cite{cho2014learning} with dimensions $h_t\in\mathbb{R}^{100}$. We consider different combinations of sentence representations: (i) \textbf{RNN}$^{w}$, (ii)  \textbf{RNN}$^{wp}$, (iii)  \textbf{RNN}$^{wl}$, and (iv) \textbf{RNN}$^{wpl}$, where $w$, $p$, $l$, correspond to the word embedding~\cite{pennington2014glove}, POS tag, and LIWC sequence representations (100 dimensions), respectively\footnote{When a sentence is represented through more than one sequence representation, we \emph{merge} the sequence representations in their respective embedding spaces.}. We train the model for 10 \emph{epochs} with a \emph{batch size} of 100, and use \emph{Adam} for optimizing our \emph{binary crossentropy} loss function. We use 70\% of data for training, 10\% for validation, and the remaining 20\% for testing.
\end{itemize}

We also used a sentiment classifier for the problem of detecting statements with biased phrasing, but the performance was too low to serve as a solid baseline. This confirms that bias detection, as a problem, differs strongly from the problem of sentiment analysis.

\subsection{Approach Learning Setup}\label{sec:learning_setup}

Here, we describe the learning setup of our two approaches: (i) RNN with attention \textbf{RNN$_a$} and (ii) RNN with hierarchical attention \textbf{RNN$_{h}$}. Similar as for the simple \textbf{RNN} baseline, we consider variations of sentence representations (see Section~\ref{subsec:representation}). For all representations, we consider an embedding space of 100 dimensions, that is, $W_{emb}\in \mathbb{R}^{k\times100}$, where $k$ is the number of entries in the respective representation space.

We use Keras with Tensorflow as a backend. We again train for 10 epochs with batch size of 100 and use 70\% of data for training, 10\% for validation, and the remaining 20\% for testing. We minimize the \emph{binary crossentropy} loss w.r.t the \emph{accuracy} metric.

We consider the following configurations for our approaches:
\begin{itemize}[leftmargin=*]
	\item \textbf{RNN}$_a$: To represent the sequences in terms of POS tags and the word function based on LIWC, we need to train the corresponding embeddings, in which case, we consider three scenarios: (i) train separately the embedding weights, (ii) share the weights amongst POS tag and LIWC representations of sentences, and (iii) share the weights amongst all three sentence representations.
	\item \textbf{RNN}$_h$: In the case of the hierarchical attention, we represent a sentence in either 2 dimensions through its word and (POS or LIWC representation), or through all its three representations. In terms of embeddings, we consider pre-trained word embeddings~\cite{pennington2014glove} or train word embeddings jointly with POS and LIWC representations together.
\end{itemize}

\section{Evaluation Results} \label{sec:eval}

In this section, we present the evaluation results and a detailed discussion. We focus on two main aspects: (i) performance in predicting if a statement contains biased language, and (ii) robustness, where we consider a real-world scenario of predicting if statements in revisions in a Wikipedia article contain biased language.

\begin{table*}[ht!]
\centering
 \resizebox{\textwidth}{!}{%
\begin{tabular}{ p{1cm}  c c c c  c c c c  c c c c cccc}
\toprule
& \multicolumn{4}{c}{\emph{\textbf{type-balanced}}} & \multicolumn{4}{c}{\textbf{\emph{featured}}} & \multicolumn{4}{c}{\textbf{\emph{cw-hard}}} & \multicolumn{4}{c}{\emph{\textbf{average}}} \\
 & \multicolumn{1}{c}{Acc} & \multicolumn{1}{c}{P} & \multicolumn{1}{c}{R} & \multicolumn{1}{c}{F1} & \multicolumn{1}{c}{Acc} & \multicolumn{1}{c}{P} & \multicolumn{1}{c}{R} & \multicolumn{1}{c}{F1} & \multicolumn{1}{c}{Acc} & \multicolumn{1}{c}{P} & \multicolumn{1}{c}{R} & \multicolumn{1}{c}{F1} & $\overline{Acc}$ & \emph{MAP} & $\overline{R}$ & $\overline{F1}$ \\
\midrule
\textbf{B1} & 0.666 & 0.669 & 0.657 & 0.663 & 0.646 & 0.650 & 0.632 & 0.641 & 0.622 & 0.626 & 0.606 & 0.616 & 0.645 & 0.648 & 0.632 & 0.640 \\
\textbf{B2} & 0.707 & 0.705 & 0.710 & 0.708 & 0.702 & 0.703 & 0.700 & 0.700 & 0.641 & 0.640 & 0.645 & 0.643  & 0.683 & 0.683 & 0.685 & 0.684\\[1ex]

\textbf{RNN}$^w$ & 0.786 & 0.805 & 0.738 & 0.770 & 0.776 &  0.788 & 0.780 & 0.784 & 0.653 & 0.668 & 0.668 & 0.668 & 0.738 & 0.754 & 0.729 & 0.741\\
\textbf{RNN}$^{wp}$ & 0.802 & \textbf{0.839} & 0.722 & 0.776 & 0.789 & 0.843 & 0.717 & 0.775 & 0.653 & 0.709 & 0.524 & 0.602 & 0.748 & 0.797 & 0.654 & 0.718\\
\textbf{RNN}$^{wl}$ & 0.779 & 0.716 & \textbf{0.869} & 0.785 & 0.794 & 0.851 & 0.717 & 0.778 & 0.651 & 0.650 & 0.715 & 0.681 & 0.741 & 0.739 & 0.767 & 0.748\\
\textbf{RNN}$^{wpl}$ & 0.773 & 0.770 & 0.762 & 0.766 & 0.771 & 0.803 & 0.738 & 0.769 & 0.648 & 0.670 & 0.639 & 0.654  & 0.731 & 0.748 & 0.713 & 0.730\\[1ex]

\textbf{RNN}$_a^w$ & 0.783 & 0.784 & 0.767 & 0.776 & 0.795 &  0.866 & 0.691 & 0.769 & 0.686 & 0.699 & 0.699 & 0.699  & 0.755 & 0.783 & 0.719 & 0.748\\
\textbf{RNN}$_a^{wp}$ & 0.803 & 0.801 & 0.794 & 0.797 & 0.818 & 0.892 & 0.715 &  0.794 & 0.681 & \textbf{0.712} & 0.647 & 0.678  & 0.767 & 0.802 & 0.719 & 0.756 \\
\textbf{RNN}$_a^{wl}$ & \textbf{0.808} & 0.814 & 0.786 & \textbf{0.800} & 0.800 & 0.809 & \textbf{0.809} & 0.809 & 0.688 &  0.697 & 0.712 & \textbf{0.705}  & 0.765 & 0.773 & \textbf{0.769} & \textbf{0.771}\\
\textbf{RNN}$_a^{wpl}$ & 0.796 & 0.820 & 0.741 & 0.778 & 0.801 & 0.860 & 0.723 & 0.785 & \textbf{0.691} & 0.710 & 0.691 &  0.700  & 0.763 & 0.797 & 0.718 & 0.754\\[1ex]

\textbf{RNN}$_h^{wp}$ & 0.796 & 0.832 & 0.714 & 0.768 & 0.803 & 0.899 & 0.649 & 0.754 & 0.664 & 0.689 & 0.644 & 0.666 & 0.754 & \textbf{0.807} & 0.669 & 0.729\\
\textbf{RNN}$_h^{wl}$ & 0.785 & 0.809 & 0.725 & 0.764 & \textbf{0.819} & \textbf{0.917} & 0.668 & 0.773 & 0.672 & 0.665 & \textbf{0.743} & 0.702 & 0.759 & 0.797 & 0.712 & 0.746\\
\textbf{RNN}$_h^{wpl}$ & 0.807 & 0.837 & 0.741 & 0.786 & 0.812 & 0.872 & 0.733 & 0.797 & 0.679 & 0.696 & 0.683 & 0.690 & \textbf{0.766} & 0.802 & 0.719 & 0.758\\
\bottomrule
\end{tabular}}
\caption{\small{Evaluation results for all competing approaches. We show the results for all three different datasets. The evaluation metrics (P/R/F1) are shown for the \emph{``biased''} class. The best scores for each metric and dataset are marked in bold.}}
\label{performance}

\vspace{-5mm}
\end{table*}

\subsection{Biased Language Detection Performance}\label{subsec:performance}

Table~\ref{performance} shows the evaluation results for all competitors and the different configurations of our approach in classifying statements if they contain biased language. The results are shown for the three different datasets, which vary only in terms of \emph{``neutral''} statements, specifically how we sample for such statements (see Section~\ref{sec:datasets}).

\textbf{Feature-Based.} We see that feature based algorithms like the baselines in \textbf{B1} and \textbf{B2} are outperformed by all RNN based approaches. This confirms our hypothesis that biased language is often introduced through multiple words or phrases that are hard to capture through word lexicons~\cite{recasens2013linguistic}.  We notice that n-gram features in \textbf{B2} provide a relative improvement of 6.8\% in terms of F1 score for the \emph{type-balanced} dataset. Similar improvements are observed for the other datasets. It is worth noting that in the case of the \emph{\textbf{cw-hard}} dataset, the performance is significantly lower when compared to the other two datasets, with a relative decrease of 10\% in terms of F1 score for the \emph{\textbf{type-balanced}} dataset. This is attributed to the difficulty in distinguishing between \emph{``biased''} and \emph{``neutral''} statements in \emph{\textbf{cw-hard}}, since neutral statements in this case contain other forms of bias such as selection, focus bias etc.

\textbf{RNN baselines.} Our main claim in this work was that language bias in statements is hard to capture through n-gram based features, and that RNN based models can better capture the inter-dependencies between words and phrases that introduce bias. Table~\ref{performance} confirms this claim. If we consider only the RNN baselines with GRU cells~\cite{cho2014learning}, the best configuration is when representing the sentence as a combination of its words and the LIWC function of a word, specifically through the concatenated embeddings of both representations. RNN$^{wl}$ achieves a relative improvement of 11\% in terms of F1 score for the \emph{\textbf{type-balanced}} dataset. Similar improvements are observed in the other two remaining datasets. In terms of precision the improvement can go well beyond 19\%, whereas in terms of recall we see an improvement of 22\%. This shows the ability of RNN based approaches to encode sequences in a statement such that only sequences which help in the classification task, respectively their information from the hidden states, are passed onto the sentence encoding (Eq (1) -- (4)), thus, making the classification much more accurate.

\textbf{Attention-based RNN.} The attention mechanism allows us to capture the importance of the specific input sequences from a sentence for the classifying task. We employ two modes of attention. First, the global attention that operates on top of the merged sentence representation RNN$_a$. Second, a hierarchical attention, which is first applied on the separate sentence representations, whereby we construct an intermediate sentence representation based on the most important input sequences, and on top of which we apply another layer of attention, and finally classify the sentence. 

We note that RNNs with hierarchical attention achieve the best performance amongst all approaches, with $P=0.917$ in the setting of RNN$_{h}^{wl}$, whereas RNN$_{a}^{wp}$ achieves $P=0.892$. This presents an improvement of over 30\% in terms of precision over the feature-based model \textbf{B2}, and 5\% improvement over the best performing RNN baseline. In terms of F1 score, RNN$_a$ achieves the best performance, due to their higher coverage of \emph{``biased''} statements. 

A direct comparison between the two modes of attention reveals that the performance is quite close. Hierarchical attention achieves overall better precision, however, at the cost of recall. Interestingly, we see that in all cases there is a gain in representing statements through the word, POS and LIWC word function representations. This shows that context (through word embeddings) and in combination with the linguistic style that is captured through POS tags and additionally the LIWC word functions can yield significant improvement over simplistic word representations. 

Over all datasets, we see that in terms of accuracy and precision RNN$_h$ perform best, whereas in terms of F1 score RNN$_a$ shows the best performance. In the next task, where we assess the robustness of our model, we pick RNN$_a^{wl}$ as it is most stable in terms of F1 across all datasets.

\begin{table}[ht!]
\centering
\begin{tabular}{ l  c c c c }
\toprule
& \textbf{Acc} & \textbf{P} & \textbf{R} & \textbf{F1}  \\
 \midrule
\textbf{type-balanced} & 0.638 & 0.609 & 0.757 & 0.675 \\
\textbf{featured} & 0.678 & 0.654 & 0.757 & 0.702 \\
\textbf{cw-hard} & 0.645 & 0.640 & 0.686 & 0.662 \\
\bottomrule
\end{tabular}
\caption{\small{Robustness results for our best performing approach and the impact of its training on the different datasets. }}
\label{table:robustness}
\vspace{-7mm}
\end{table}

\subsection{Robustness}\label{sec:robust}

For a large variety of tasks, an important concern is how well do trained model on controlled settings perform in real-world scenarios? To this end, we assess the robustness of our approach by considering statements coming from a \emph{controversial}\footnote{\url{https://en.wikipedia.org/wiki/Wikipedia:List_of_controversial_issues}} Wikipedia article \texttt{Abortion}\footnote{\url{https://en.wikipedia.org/wiki/Abortion}}. This article serves only to demonstrate how well our best performing model RNN$_a^{wl}$ pre-trained in the previous three datasets would perform in correctly classifying statements in this article that contain biased language.

From the entire revision history of the \texttt{Abortion} article, we extract revisions that contain \emph{POV} quality tags, and thus, extract all statements that have been deleted or modified. There are different reasons why editors delete or modify statements, as indicated by editor comments. Examples apart from POV issues are statements considered to be \emph{irrelevant} or \emph{unimportant}, statements that are \emph{not supported by a source}, or \emph{vandalism}. This resulted in 10,243 statements, from which we sample 100 and annotate them through crowdsourcing, similar as in Section~\ref{sec:data_collection}. The annotated dataset contains 52 statements labeled as biased and 48 statements labeled as neutral. The high number of statements labeled as biased is not surprising given the controversial topic of the article.

Table \ref{table:robustness} shows the performance of the best performing model RNN$_a^{wl}$ pre-trained on the datasets in Section~\ref{sec:datasets}, and evaluated on the \emph{robustness data}. The performance of the model trained on the \emph{\textbf{type-balanced}} and the \emph{\textbf{featured}} datasets is stable with F1 scores of 67.5\% and 70.2\%. 

Similarly, as in Table~\ref{performance}, we see a lower performance in terms of F1 score for the model trained on the \emph{\textbf{cw-hard}} dataset. Table~\ref{table:robustness} shows that the classifiers are \emph{robust} and \emph{generalize} well over instances that are very different from their original train set. Additionally, this shows that even if we employ our approach in a real-world scenario to flag highly voluminous and unclean statements, we can detect with reasonably good performance statements that contain biased language.

\vspace{-1mm}

\section{Conclusion and Future Work}

In this paper we presented an RNN based approach for classifying statements that contain biased language. We focused on the case of biased phrasing, that is, statements in which words or phrases are \emph{inflammatory} or \emph{partial}. We showed that RNN models are superior in performance when compared to feature-based models, and are able to capture the important words and phrases that introduce bias in a statement. Furthermore, we show that encoding the statements based on different representations such as words, POS, and LIWC word function, through which we capture \emph{context}, \emph{style}, and \emph{psychological} and \emph{sociological} functions of words, we can predict with high accuracy statements that contain biased language. 

Finally, we show that with employing attention mechanisms (both global and hierarchical) we can further improve the performance of our approach, by identifying salient sequences and additionally providing means of interpreting and uncovering different forms of biased language. We are able to predict with a very high precision of $91.7\%$ , thus, providing a highly significant relative improvement over competitors with more than 30\% in terms of precision.

As future work we foresee analyzing the different forms of bias such as selection bias, and bias introduced due to the demographics of the underlying editor population in Wikipedia. 
\vfill

\paragraph*{Acknowledgments} This work is funded by the ERC Advanced Grant ALEXANDRIA (grant no. 339233), DESIR (grant no. 731081), H2020 AFEL project (grant no. 687916), and SimpleML (grant no. 01IS18054).

\balance



\begin{thebibliography}{00}

%%% the Web address.

\ifx \showCODEN    \undefined \def \showCODEN     #1{\unskip}     \fi
\ifx \showDOI      \undefined \def \showDOI       #1{#1}\fi
\ifx \showISBNx    \undefined \def \showISBNx     #1{\unskip}     \fi
\ifx \showISBNxiii \undefined \def \showISBNxiii  #1{\unskip}     \fi
\ifx \showISSN     \undefined \def \showISSN      #1{\unskip}     \fi
\ifx \showLCCN     \undefined \def \showLCCN      #1{\unskip}     \fi
\ifx \shownote     \undefined \def \shownote      #1{#1}          \fi
\ifx \showarticletitle \undefined \def \showarticletitle #1{#1}   \fi
\ifx \showURL      \undefined \def \showURL       {\relax}        \fi
% invisible to TeX
\providecommand\bibfield[2]{#2}
\providecommand\bibinfo[2]{#2}
\providecommand\natexlab[1]{#1}
\providecommand\showeprint[2][]{arXiv:#2}

\bibitem[\protect\citeauthoryear{Bahdanau, Cho, and Bengio}{Bahdanau
  et~al\mbox{.}}{2014}]%
        {bahdanau2014neural}
\bibfield{author}{\bibinfo{person}{Dzmitry Bahdanau},
  \bibinfo{person}{Kyunghyun Cho}, {and} \bibinfo{person}{Yoshua Bengio}.}
  \bibinfo{year}{2014}\natexlab{}.
\newblock \showarticletitle{Neural machine translation by jointly learning to
  align and translate}.
\newblock \bibinfo{journal}{{\em arXiv preprint arXiv:1409.0473\/}}
  (\bibinfo{year}{2014}).
\newblock


\bibitem[\protect\citeauthoryear{Baumer, Elovic, Qin, Polletta, and Gay}{Baumer
  et~al\mbox{.}}{2015}]%
        {baumer2015testing}
\bibfield{author}{\bibinfo{person}{Eric Baumer}, \bibinfo{person}{Elisha
  Elovic}, \bibinfo{person}{Ying Qin}, \bibinfo{person}{Francesca Polletta},
  {and} \bibinfo{person}{Geri Gay}.} \bibinfo{year}{2015}\natexlab{}.
\newblock \showarticletitle{Testing and comparing computational approaches for
  identifying the language of framing in political news}. In
  \bibinfo{booktitle}{{\em Proceedings of the 2015 Conference of the North
  American Chapter of the Association for Computational Linguistics: Human
  Language Technologies}}. \bibinfo{pages}{1472--1482}.
\newblock


\bibitem[\protect\citeauthoryear{Biber}{Biber}{1991}]%
        {biber1991variation}
\bibfield{author}{\bibinfo{person}{Douglas Biber}.}
  \bibinfo{year}{1991}\natexlab{}.
\newblock \bibinfo{booktitle}{{\em Variation across speech and writing}}.
\newblock \bibinfo{publisher}{Cambridge University Press}.
\newblock


\bibitem[\protect\citeauthoryear{Bourgeois, Rappaz, and Aberer}{Bourgeois
  et~al\mbox{.}}{2018}]%
        {bourgeois2018selection}
\bibfield{author}{\bibinfo{person}{Dylan Bourgeois},
  \bibinfo{person}{J{\'e}r{\'e}mie Rappaz}, {and} \bibinfo{person}{Karl
  Aberer}.} \bibinfo{year}{2018}\natexlab{}.
\newblock \showarticletitle{Selection Bias in News Coverage: Learning it,
  Fighting it}. In \bibinfo{booktitle}{{\em The International World Wide Web
  Conference 2018}}.
\newblock


\bibitem[\protect\citeauthoryear{Brown, Gilman, et~al\mbox{.}}{Brown
  et~al\mbox{.}}{1960}]%
        {brown1960pronouns}
\bibfield{author}{\bibinfo{person}{Roger Brown}, \bibinfo{person}{Albert
  Gilman}, {et~al\mbox{.}}} \bibinfo{year}{1960}\natexlab{}.
\newblock \showarticletitle{The pronouns of power and solidarity}.
\newblock  (\bibinfo{year}{1960}).
\newblock


\bibitem[\protect\citeauthoryear{Callahan and Herring}{Callahan and
  Herring}{2011}]%
        {callahan2011cultural}
\bibfield{author}{\bibinfo{person}{Ewa~S Callahan} {and}
  \bibinfo{person}{Susan~C Herring}.} \bibinfo{year}{2011}\natexlab{}.
\newblock \showarticletitle{Cultural bias in Wikipedia content on famous
  persons}.
\newblock \bibinfo{journal}{{\em JASIST\/}} \bibinfo{volume}{62},
  \bibinfo{number}{10} (\bibinfo{year}{2011}).
\newblock


\bibitem[\protect\citeauthoryear{Cho, Van~Merri{\"e}nboer, Gulcehre, Bahdanau,
  Bougares, Schwenk, and Bengio}{Cho et~al\mbox{.}}{2014}]%
        {cho2014learning}
\bibfield{author}{\bibinfo{person}{Kyunghyun Cho}, \bibinfo{person}{Bart
  Van~Merri{\"e}nboer}, \bibinfo{person}{Caglar Gulcehre},
  \bibinfo{person}{Dzmitry Bahdanau}, \bibinfo{person}{Fethi Bougares},
  \bibinfo{person}{Holger Schwenk}, {and} \bibinfo{person}{Yoshua Bengio}.}
  \bibinfo{year}{2014}\natexlab{}.
\newblock \showarticletitle{Learning phrase representations using RNN
  encoder-decoder for statistical machine translation}.
\newblock \bibinfo{journal}{{\em arXiv preprint arXiv:1406.1078\/}}
  (\bibinfo{year}{2014}).
\newblock


\bibitem[\protect\citeauthoryear{Fetahu, Anand, and Anand}{Fetahu
  et~al\mbox{.}}{2015}]%
        {DBLP:conf/websci/FetahuAA15}
\bibfield{author}{\bibinfo{person}{Besnik Fetahu}, \bibinfo{person}{Abhijit
  Anand}, {and} \bibinfo{person}{Avishek Anand}.}
  \bibinfo{year}{2015}\natexlab{}.
\newblock \showarticletitle{How much is Wikipedia Lagging Behind News?}. In
  \bibinfo{booktitle}{{\em Proceedings of the {ACM} Web Science Conference,
  WebSci 2015, Oxford, United Kingdom, June 28 - July 1, 2015}}.
  \bibinfo{pages}{28:1--28:9}.
\newblock
\showDOI{%
\url{https://doi.org/10.1145/2786451.2786460}}


\bibitem[\protect\citeauthoryear{Fetahu, Markert, Nejdl, and Anand}{Fetahu
  et~al\mbox{.}}{2016}]%
        {DBLP:conf/cikm/FetahuMNA16}
\bibfield{author}{\bibinfo{person}{Besnik Fetahu}, \bibinfo{person}{Katja
  Markert}, \bibinfo{person}{Wolfgang Nejdl}, {and} \bibinfo{person}{Avishek
  Anand}.} \bibinfo{year}{2016}\natexlab{}.
\newblock \showarticletitle{Finding News Citations for Wikipedia}. In
  \bibinfo{booktitle}{{\em Proceedings of the 25th {ACM} International
  Conference on Information and Knowledge Management, {CIKM} 2016,
  Indianapolis, IN, USA, October 24-28, 2016}}. \bibinfo{pages}{337--346}.
\newblock
\showDOI{%
\url{https://doi.org/10.1145/2983323.2983808}}


\bibitem[\protect\citeauthoryear{Fowler}{Fowler}{2013}]%
        {fowler2013language}
\bibfield{author}{\bibinfo{person}{Roger Fowler}.}
  \bibinfo{year}{2013}\natexlab{}.
\newblock \bibinfo{booktitle}{{\em Language in the News: Discourse and Ideology
  in the Press}}.
\newblock \bibinfo{publisher}{Routledge}.
\newblock


\bibitem[\protect\citeauthoryear{Gentzkow and Shapiro}{Gentzkow and
  Shapiro}{2010}]%
        {gentzkow2010drives}
\bibfield{author}{\bibinfo{person}{Matthew Gentzkow} {and}
  \bibinfo{person}{Jesse~M Shapiro}.} \bibinfo{year}{2010}\natexlab{}.
\newblock \showarticletitle{What drives media slant? Evidence from US daily
  newspapers}.
\newblock \bibinfo{journal}{{\em Econometrica\/}} \bibinfo{volume}{78},
  \bibinfo{number}{1} (\bibinfo{year}{2010}).
\newblock


\bibitem[\protect\citeauthoryear{Greenstein and Zhu}{Greenstein and
  Zhu}{2012}]%
        {greenstein2012wikipedia}
\bibfield{author}{\bibinfo{person}{Shane Greenstein} {and}
  \bibinfo{person}{Feng Zhu}.} \bibinfo{year}{2012}\natexlab{}.
\newblock \showarticletitle{Is Wikipedia Biased?}
\newblock \bibinfo{journal}{{\em The American economic review\/}}
  \bibinfo{volume}{102}, \bibinfo{number}{3} (\bibinfo{year}{2012}),
  \bibinfo{pages}{343--348}.
\newblock


\bibitem[\protect\citeauthoryear{Halliday}{Halliday}{1970}]%
        {halliday1970language}
\bibfield{author}{\bibinfo{person}{Michael~AK Halliday}.}
  \bibinfo{year}{1970}\natexlab{}.
\newblock \showarticletitle{Language structure and language function}.
\newblock \bibinfo{journal}{{\em New horizons in linguistics\/}}
  \bibinfo{volume}{1} (\bibinfo{year}{1970}), \bibinfo{pages}{140--165}.
\newblock


\bibitem[\protect\citeauthoryear{Hochreiter and Schmidhuber}{Hochreiter and
  Schmidhuber}{1997}]%
        {hochreiter1997long}
\bibfield{author}{\bibinfo{person}{Sepp Hochreiter} {and}
  \bibinfo{person}{J{\"u}rgen Schmidhuber}.} \bibinfo{year}{1997}\natexlab{}.
\newblock \showarticletitle{Long short-term memory}.
\newblock \bibinfo{journal}{{\em Neural computation\/}} \bibinfo{volume}{9},
  \bibinfo{number}{8} (\bibinfo{year}{1997}), \bibinfo{pages}{1735--1780}.
\newblock


\bibitem[\protect\citeauthoryear{Hube and Fetahu}{Hube and Fetahu}{2018}]%
        {hube2018detecting}
\bibfield{author}{\bibinfo{person}{Christoph Hube} {and}
  \bibinfo{person}{Besnik Fetahu}.} \bibinfo{year}{2018}\natexlab{}.
\newblock \showarticletitle{Detecting Biased Statements in Wikipedia}. In
  \bibinfo{booktitle}{{\em Companion of the The Web Conference 2018 on The Web
  Conference 2018}}. International World Wide Web Conferences Steering
  Committee, \bibinfo{pages}{1779--1786}.
\newblock


\bibitem[\protect\citeauthoryear{Iyyer, Enns, Boyd-Graber, and Resnik}{Iyyer
  et~al\mbox{.}}{2014}]%
        {iyyer2014political}
\bibfield{author}{\bibinfo{person}{Mohit Iyyer}, \bibinfo{person}{Peter Enns},
  \bibinfo{person}{Jordan Boyd-Graber}, {and} \bibinfo{person}{Philip Resnik}.}
  \bibinfo{year}{2014}\natexlab{}.
\newblock \showarticletitle{Political ideology detection using recursive neural
  networks}. In \bibinfo{booktitle}{{\em Proceedings of the Association for
  Computational Linguistics}}. \bibinfo{pages}{1--11}.
\newblock


\bibitem[\protect\citeauthoryear{Lahoti, Garimella, and Gionis}{Lahoti
  et~al\mbox{.}}{2018}]%
        {lahoti2017Joint}
\bibfield{author}{\bibinfo{person}{Preethi Lahoti}, \bibinfo{person}{Kiran
  Garimella}, {and} \bibinfo{person}{Aristides Gionis}.}
  \bibinfo{year}{2018}\natexlab{}.
\newblock \showarticletitle{Joint Non-negative Matrix Factorization for
  Learning Ideological Leaning on Twitter}. In \bibinfo{booktitle}{{\em
  Proceedings of the Eleventh ACM International Conference on Web Search and
  Data Mining}} {\em (\bibinfo{series}{WSDM '18})}. \bibinfo{publisher}{ACM},
  \bibinfo{address}{New York, NY, USA}, \bibinfo{pages}{351--359}.
\newblock
\showISBNx{978-1-4503-5581-0}
\showDOI{%
\url{https://doi.org/10.1145/3159652.3159669}}


\bibitem[\protect\citeauthoryear{Luong, Pham, and Manning}{Luong
  et~al\mbox{.}}{2015}]%
        {luong2015effective}
\bibfield{author}{\bibinfo{person}{Minh-Thang Luong}, \bibinfo{person}{Hieu
  Pham}, {and} \bibinfo{person}{Christopher~D Manning}.}
  \bibinfo{year}{2015}\natexlab{}.
\newblock \showarticletitle{Effective approaches to attention-based neural
  machine translation}.
\newblock \bibinfo{journal}{{\em arXiv preprint arXiv:1508.04025\/}}
  (\bibinfo{year}{2015}).
\newblock


\bibitem[\protect\citeauthoryear{Martin}{Martin}{2017}]%
        {martin2017persistent}
\bibfield{author}{\bibinfo{person}{Brian Martin}.}
  \bibinfo{year}{2017}\natexlab{}.
\newblock \showarticletitle{Persistent Bias on Wikipedia: Methods and
  Responses}.
\newblock \bibinfo{journal}{{\em Social Science Computer Review\/}}
  (\bibinfo{year}{2017}), \bibinfo{pages}{0894439317715434}.
\newblock


\bibitem[\protect\citeauthoryear{Pennebaker, Francis, and Booth}{Pennebaker
  et~al\mbox{.}}{2001}]%
        {pennebaker2001linguistic}
\bibfield{author}{\bibinfo{person}{James~W Pennebaker},
  \bibinfo{person}{Martha~E Francis}, {and} \bibinfo{person}{Roger~J Booth}.}
  \bibinfo{year}{2001}\natexlab{}.
\newblock \showarticletitle{Linguistic inquiry and word count: LIWC 2001}.
\newblock \bibinfo{journal}{{\em Mahway: Lawrence Erlbaum Associates\/}}
  \bibinfo{volume}{71}, \bibinfo{number}{2001} (\bibinfo{year}{2001}),
  \bibinfo{pages}{2001}.
\newblock


\bibitem[\protect\citeauthoryear{Pennington, Socher, and Manning}{Pennington
  et~al\mbox{.}}{2014}]%
        {pennington2014glove}
\bibfield{author}{\bibinfo{person}{Jeffrey Pennington},
  \bibinfo{person}{Richard Socher}, {and} \bibinfo{person}{Christopher
  Manning}.} \bibinfo{year}{2014}\natexlab{}.
\newblock \showarticletitle{Glove: Global vectors for word representation}. In
  \bibinfo{booktitle}{{\em Proceedings of the 2014 conference on empirical
  methods in natural language processing (EMNLP)}}.
  \bibinfo{pages}{1532--1543}.
\newblock


\bibitem[\protect\citeauthoryear{Rashkin, Choi, Jang, Volkova, and
  Choi}{Rashkin et~al\mbox{.}}{2017}]%
        {rashkin2017truth}
\bibfield{author}{\bibinfo{person}{Hannah Rashkin}, \bibinfo{person}{Eunsol
  Choi}, \bibinfo{person}{Jin~Yea Jang}, \bibinfo{person}{Svitlana Volkova},
  {and} \bibinfo{person}{Yejin Choi}.} \bibinfo{year}{2017}\natexlab{}.
\newblock \showarticletitle{Truth of varying shades: Analyzing language in fake
  news and political fact-checking}. In \bibinfo{booktitle}{{\em Proceedings of
  the 2017 Conference on Empirical Methods in Natural Language Processing}}.
  \bibinfo{pages}{2931--2937}.
\newblock


\bibitem[\protect\citeauthoryear{Recasens, Danescu-Niculescu-Mizil, and
  Jurafsky}{Recasens et~al\mbox{.}}{2013}]%
        {recasens2013linguistic}
\bibfield{author}{\bibinfo{person}{Marta Recasens}, \bibinfo{person}{Cristian
  Danescu-Niculescu-Mizil}, {and} \bibinfo{person}{Dan Jurafsky}.}
  \bibinfo{year}{2013}\natexlab{}.
\newblock \showarticletitle{Linguistic Models for Analyzing and Detecting
  Biased Language.}. In \bibinfo{booktitle}{{\em ACL (1)}}.
  \bibinfo{pages}{1650--1659}.
\newblock


\bibitem[\protect\citeauthoryear{Romaine et~al\mbox{.}}{Romaine
  et~al\mbox{.}}{2000}]%
        {romaine2000language}
\bibfield{author}{\bibinfo{person}{Suzanne Romaine} {et~al\mbox{.}}}
  \bibinfo{year}{2000}\natexlab{}.
\newblock \bibinfo{booktitle}{{\em Language in society: An introduction to
  sociolinguistics}}.
\newblock \bibinfo{publisher}{Oxford University Press}.
\newblock


\bibitem[\protect\citeauthoryear{Tsur, Calacci, and Lazer}{Tsur
  et~al\mbox{.}}{2015}]%
        {tsur2015frame}
\bibfield{author}{\bibinfo{person}{Oren Tsur}, \bibinfo{person}{Dan Calacci},
  {and} \bibinfo{person}{David Lazer}.} \bibinfo{year}{2015}\natexlab{}.
\newblock \showarticletitle{A frame of mind: Using statistical models for
  detection of framing and agenda setting campaigns}. In
  \bibinfo{booktitle}{{\em Proceedings of the 53rd Annual Meeting of the
  Association for Computational Linguistics and the 7th International Joint
  Conference on Natural Language Processing (Volume 1: Long Papers)}},
  Vol.~\bibinfo{volume}{1}. \bibinfo{pages}{1629--1638}.
\newblock


\bibitem[\protect\citeauthoryear{Wagner, Garcia, Jadidi, and Strohmaier}{Wagner
  et~al\mbox{.}}{2015}]%
        {wagner2015s}
\bibfield{author}{\bibinfo{person}{Claudia Wagner}, \bibinfo{person}{David
  Garcia}, \bibinfo{person}{Mohsen Jadidi}, {and} \bibinfo{person}{Markus
  Strohmaier}.} \bibinfo{year}{2015}\natexlab{}.
\newblock \showarticletitle{It's a man's wikipedia? assessing gender inequality
  in an online encyclopedia}.
\newblock \bibinfo{journal}{{\em arXiv preprint arXiv:1501.06307\/}}
  (\bibinfo{year}{2015}).
\newblock


\bibitem[\protect\citeauthoryear{Yang, Yang, Dyer, He, Smola, and Hovy}{Yang
  et~al\mbox{.}}{2016}]%
        {yang2016hierarchical}
\bibfield{author}{\bibinfo{person}{Zichao Yang}, \bibinfo{person}{Diyi Yang},
  \bibinfo{person}{Chris Dyer}, \bibinfo{person}{Xiaodong He},
  \bibinfo{person}{Alex Smola}, {and} \bibinfo{person}{Eduard Hovy}.}
  \bibinfo{year}{2016}\natexlab{}.
\newblock \showarticletitle{Hierarchical attention networks for document
  classification}. In \bibinfo{booktitle}{{\em Proceedings of the 2016
  Conference of the North American Chapter of the Association for Computational
  Linguistics: Human Language Technologies}}. \bibinfo{pages}{1480--1489}.
\newblock


\bibitem[\protect\citeauthoryear{Yano, Resnik, and Smith}{Yano
  et~al\mbox{.}}{2010}]%
        {yano2010shedding}
\bibfield{author}{\bibinfo{person}{Tae Yano}, \bibinfo{person}{Philip Resnik},
  {and} \bibinfo{person}{Noah~A Smith}.} \bibinfo{year}{2010}\natexlab{}.
\newblock \showarticletitle{Shedding (a thousand points of) light on biased
  language}. In \bibinfo{booktitle}{{\em Proceedings of the NAACL HLT 2010
  Workshop on Creating Speech and Language Data with Amazon's Mechanical
  Turk}}. Association for Computational Linguistics, \bibinfo{pages}{152--158}.
\newblock


\end{thebibliography}
\end{document}